\documentclass[letterpaper, 10 pt, conference]{ieeeconf}  
\usepackage{graphicx}
\usepackage{multirow}
\usepackage{color}
\usepackage{amsmath, amssymb, amscd, amsfonts} 
\let\labelindent\relax
\usepackage{enumitem}
\usepackage{graphics} 
\usepackage{graphicx}
\usepackage{hyperref}
\usepackage{tabularx}
\usepackage{multirow}
\usepackage{vcell}
\usepackage{colortbl}
\usepackage{booktabs}
\usepackage{cite}
\usepackage{float}
\usepackage{subfig}
\usepackage{caption}
\usepackage{algorithm}
\usepackage{algorithmic}
\usepackage{bbding}
\usepackage{array}
\usepackage{makecell}
\usepackage{lipsum}
\usepackage{tikz}
\usepackage{url}
\usepackage{tabularray}

\IEEEoverridecommandlockouts                              

\title{\LARGE \bf
RMP-YOLO: A Robust Motion Predictor for Partially Observable Scenarios even if You Only Look Once

\author{
    Jiawei Sun$^{1}$, Jiahui Li$^{1}$, Tingchen Liu$^{1}$, Chengran Yuan$^{1}$, Shuo Sun$^{1}$, \\ Zefan Huang$^{1}$, Anthony Wong$^{2}$,
    Keng Peng Tee$^{2}$ and Marcelo H. Ang Jr$^{1}$.
\thanks{$^{1}$Jiawei Sun, Jiahui Li,  Tingchen Liu, Chengran Yuan, Shuo Sun, Zefan Huang and Marcelo H. Ang Jr. are with the Department of Mechanical
Engineering, National University of Singapore, Singapore 119077 (e-mail: \{sunjiawei, e1373481, e1010862, chengran.yuan, shuo.sun, huangzefan
\}@u.nus.edu; mpeangh@nus.edu.sg).}
\thanks{$^{2}$Keng Peng Tee, Anthony Wong
are with Moovita Pte Ltd, Singapore, 599489 (e-mail:{ anthonywong,
kptee}@moovita.com).}
\thanks{This work was supported in part by MooVita Pte. Ltd, Yinson and the National Research Foundation, Prime Minister’s Office, Singapore, through the CREATE Programme, as well as by the Singapore-MIT Alliance for Research and Technology (SMART) Mens, Manus, and Machina (M3S) Interdisciplinary Research Group (IRG).}
}}

\begin{document}

\maketitle
\thispagestyle{empty}
\pagestyle{empty}

\begin{abstract}
We introduce RMP-YOLO, a unified framework designed to provide robust motion predictions even with incomplete input data. 
Our key insight stems from the observation that complete and reliable historical trajectory data plays a pivotal role in ensuring accurate motion prediction. Therefore, we propose a new paradigm that prioritizes the reconstruction of intact historical trajectories before feeding them into the prediction modules. Our approach introduces a novel scene tokenization module to enhance the extraction and fusion of spatial and temporal features. Following this, our proposed recovery module reconstructs agents' incomplete historical trajectories by leveraging local map topology and interactions with nearby agents. The reconstructed, clean historical data is then integrated into the downstream prediction modules. Our framework is able to effectively handle missing data of varying lengths and remains robust against observation noise while maintaining high prediction accuracy. Furthermore, our recovery module is compatible with existing prediction models, ensuring seamless integration. Extensive experiments validate the effectiveness of our approach, and deployment in real-world autonomous vehicles confirms its practical utility. In the 2024 Waymo Motion Prediction Competition, our method, RMP-YOLO, achieves state-of-the-art performance, securing third place. Our code is open-source at \href{https://github.com/ggosjw/RMP-YOLO}{https://github.com/ggosjw/RMP-YOLO}.

\end{abstract}

\section{INTRODUCTION}
Motion prediction is an essential module in the autonomous driving system. Recent motion prediction methods mostly adopt a learning-based approach that relies on the High-Definition (HD) lane map and the other agents' observed historical trajectories from the upstream tracking module as inputs to predict the other agents' future trajectories. The release of several large-scale driving datasets, including the Waymo Open Dataset\cite{ettinger2021largescaleinteractivemotion,  waymo_motion_prediction}, Argoverse 1\cite{chang2019argoverse, argoverse}, and Argoverse 2\cite{ettinger2021large, argoverse2}, has tremendously accelerated the development in learning-based prediction models. These datasets provide high-quality driving data collected in the real world, benchmarks, and public competitions to gauge the model's performance. 

\begin{figure}[t]
    \centering
    \includegraphics[width=8cm]{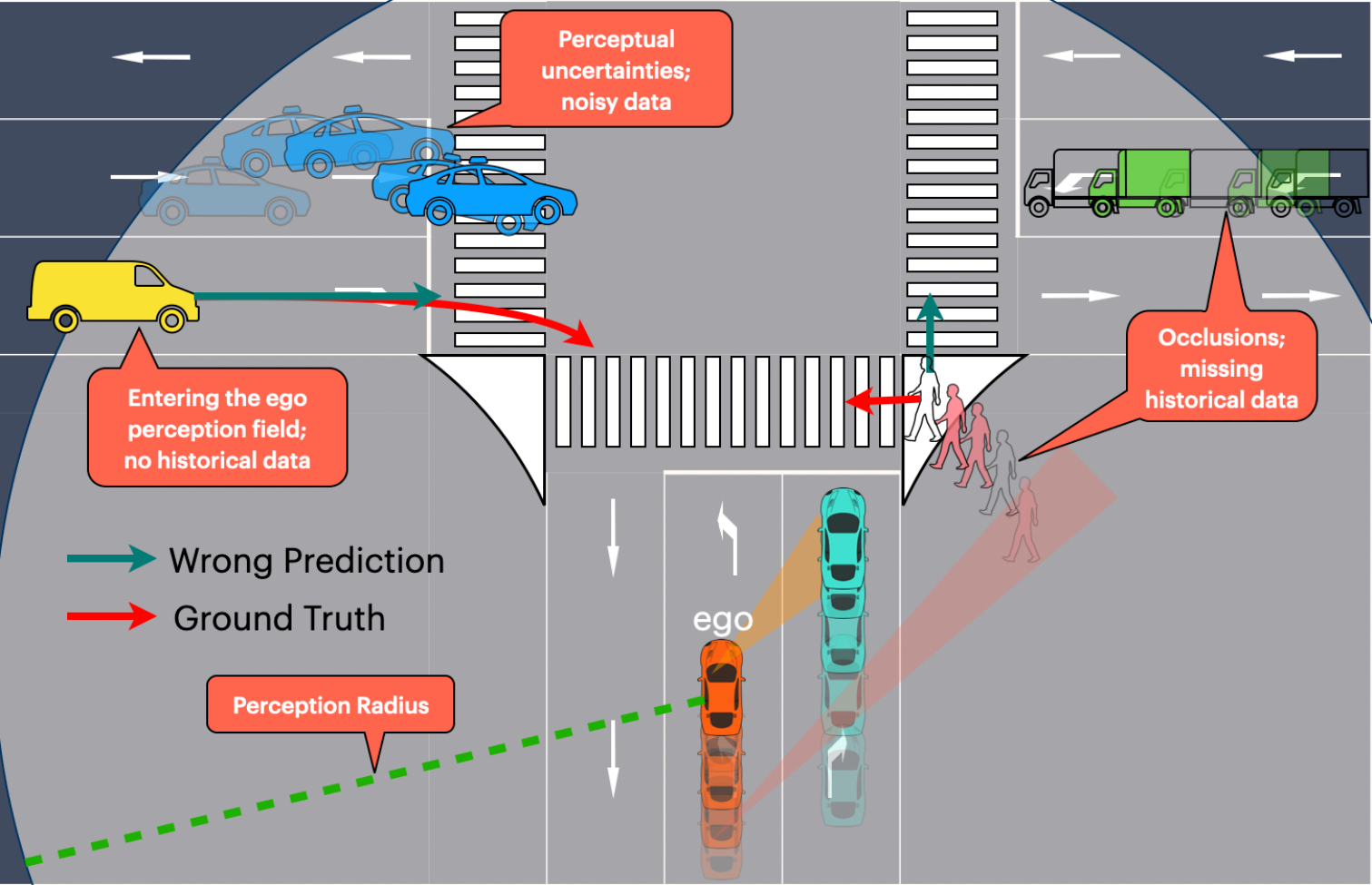} 
    \caption{Example of a partially observable prediction scenario. The cyan car next to the red ego vehicle has complete trajectories, while the green truck and coral-colored vehicle are occluded. The blue car's data is noisy, and the yellow truck has been tracked for only one timestamp. These challenges make prediction especially difficult.}
    \label{fig:cover}
\end{figure}

However, there exists a huge discrepancy between the selected agents in public datasets and those in the real world, which is caused by imperfections in the agents' historical trajectories, as illustrated in Fig. \ref{fig:cover}. In reality, all the agents' historical trajectories are observed from the ego vehicle's perspective and processed by the upstream modules. Thus, unavoidably, these observed historical trajectories contain several imperfections, including 1) an insufficient number of history time steps over the period when every agent was first detected and tracked, 2) missing time steps due to out-of-view, occlusion, and ID-switch cases, and 3) the omnipresent perceptual uncertainties (noise) at all time steps. While perceptual noise is generally retained, existing public benchmarks and competitions tend to focus only on carefully selected agents with complete and high-quality historical trajectories. Non-selected agents with flawed observations are often excluded from both the supervision during training and the evaluation of model performance. 

\begin{figure}[h]
    \centering
    \includegraphics[width=1\linewidth]{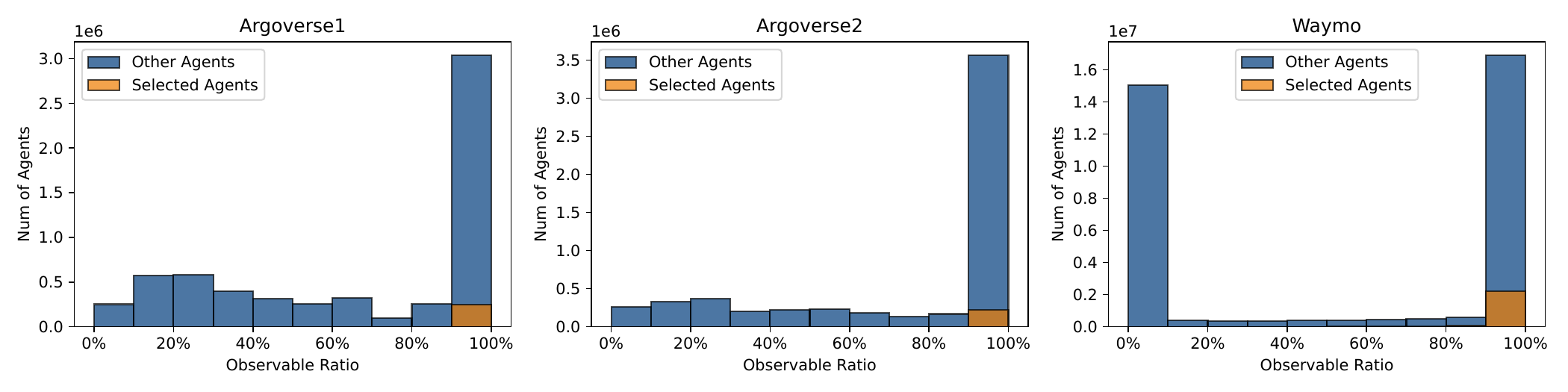} 
    \caption{Distributions of selected agents vs. non-selected agents in Waymo, Argoverse 1\&2 datasets.}
    \label{fig:distribution}
\end{figure}

After analyzing the data distribution of the agents' trajectories in popular datasets, it was found that 100\% of the selected agents in Argoverse 1 \& 2 datasets come with complete historical trajectories, while in the Waymo dataset, the selected agents have an average of 97\% observable historical trajectories. This can be a serious deviation (as shown in Fig. \ref{fig:distribution}) from the real-world distribution as all agents do not come with complete historical data in real life. As a result, methods trained and evaluated solely on clean datasets may perform well on leaderboards, able to predict accurately when complete historical trajectories are provided. However, without the luxury of having near-perfect historical trajectories, these methods often experience significant performance degradation due to this domain shift (see Fig.\ref{fig:all_comparison}) and produce erroneous predictions in real deployments. To bridge the gap between training with sanitized data and the deployment in real life with noisy, imperfect data, it is critical to develop models that are robust to data imperfections.

In this paper, we propose the RMP-YOLO framework, which offers three key advantages simultaneously: it maintains excellent prediction performance, effectively handles varying distinct missing input lengths, and remains robust against noisy inputs. Remarkably, it can still provide accurate predictions even when the target has just entered the ego agent's perceptive field with only a single timestamp of valid data. Our key insight is that since complete and noise-free historical trajectories are crucial for accurate motion prediction, it makes sense to prioritize their reconstruction first.
By leveraging local topological map structures and the relationships between nearby agents, even a limited number of observed timesteps, sometimes just a single valid frame, can provide valuable cues for the reconstruction. These local relationships enable us to more accurately infer the complete historical trajectory, thereby improving predictions for future motion. To achieve this, we introduce a simple MLP-based recovery module combined with one layer of local attention transformer to reconstruct the historical trajectories. Our contributions are summarized as follows:
\begin{itemize} 
    \item[1)] We introduce RMP-YOLO, a novel framework that prioritizes reconstructing agents' incomplete historical trajectories by leveraging local map topology and agent interactions. This reconstruction process effectively handles varying lengths of missing data, ensuring robustness against noise and incomplete observations. 
    \item[2)] The recovery module we propose is simple and lightweight. Moreover, it integrates seamlessly with existing motion prediction models, enhancing their robustness without the need for extensive modifications.
    \item[3)] Our method won third place in the 2024 Waymo Motion Prediction Competition. We deploy our algorithms on real vehicles to validate the effectiveness of our methods.  
\end{itemize}

\section{Related Works}
\subsection{Motion Prediction}

Recent advancements\cite{shi2023MTR,Zhou_2023_qc,shi2024mtr++,lan2023SEPT,Feng_2023_macformer,li2024admaccelerateddiffusionmodel} in multi-agent motion prediction have introduced several innovative approaches aimed at improving prediction accuracy and efficiency. Among these, two of the most influential prediction paradigms are MTR \cite{shi2023MTR} and QCNet \cite{Zhou_2023_qc}. MTR addresses the trade-off between goal-based prediction\cite{gu2021densetnt,zhao2020tnt,fang2021tpnet,WangGanet} and direct regression-based\cite{zhang2024simple,Feng_2023_macformer,fang2021tpnet,varadarajan2021multipath++,ngiam2022scenetransformer} prediction, by its proposed motion query pairs, while QCNet proposes a symmetric\cite{JiaHDGT}, translation-invarnt\cite{Hivt} input representation which enables reusable stream prediction.
Plenty of derivative works have expanded on these foundational approaches, including EDA \cite{lin2023eda}, MTR++ \cite{shi2024mtr++}, ControlMTR \cite{sun2024controlmtrcontrolguidedmotiontransformer}, SIMPL \cite{zhang2024simple}, MGTR \cite{gan2024mgtrmultigranulartransformermotion}, and others \cite{zhou2023qcnext,zheng2024LLM_AUGUMENT,tang2024hpnet,mu2024mostmultimodalityscenetokenization,zhang2023hptr}.
However, a common limitation shared by these methods is that they are typically trained on nearly complete input trajectories. As a result, their performance may degrade when confronted with imperfect or incomplete data during inference (see Fig.\ref{fig:all_comparison}). To address this, we propose the RMP-YOLO framework, building upon MTR, to effectively handle input data imperfections while maintaining high prediction accuracy. 


\subsection{Imperfect Data Recovery}
\begin{figure}[h]
    \centering
    \includegraphics[width=0.8\linewidth]{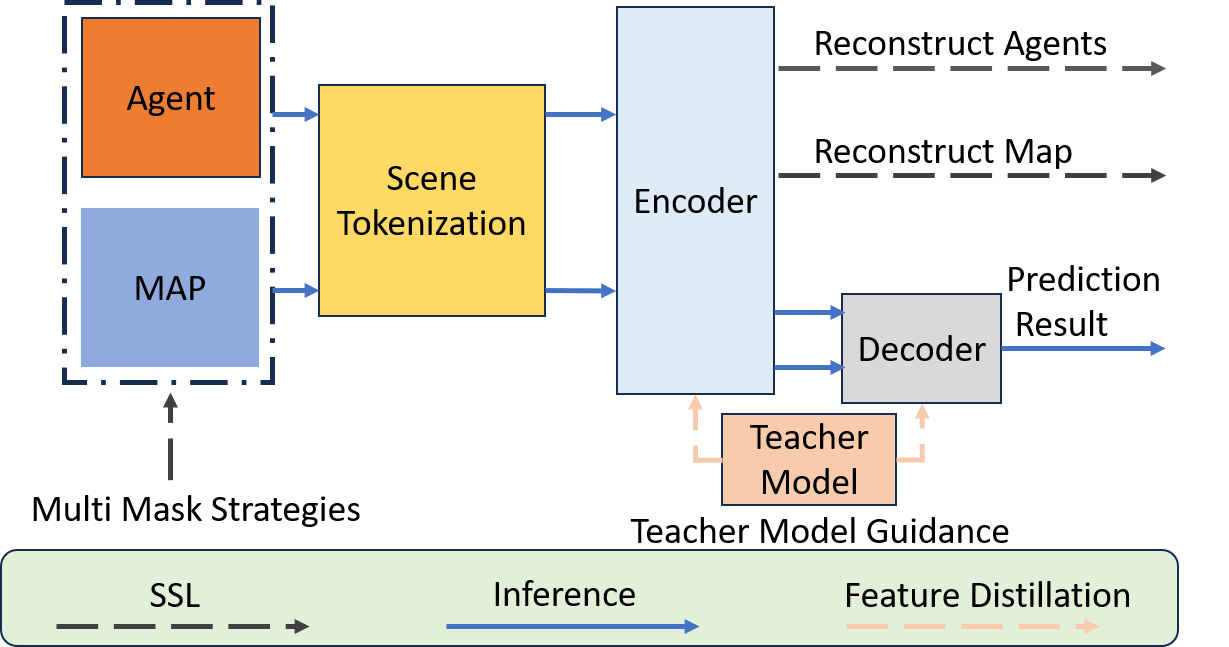}
    \caption{Generalized pipeline for previous methods addressing partially observable historical trajectories.}
    \label{fig:general-pipline}
\end{figure}

Previous works tackling imperfect data generally focus on either incomplete input trajectories\cite{POP_RAL,random_mask_pretrain,xu2023uncoveringmissingpatternunified,monti2022observationsenoughknowledgedistillation} or noisy input data\cite{adversarialbackdoorattacknaturalistic,On_adversarial_robustness}. To handle incomplete input data, self-supervised learning and feature distillation are commonly employed techniques. A generalized pipeline for this kind of methods is illustrated in Fig.\ref{fig:general-pipline}. SSL-Lanes\cite{bhattacharyya2022ssllane} and Forecast-MAE\cite{forecastmae} are pioneering works utilizing SSL techniques for vehicle motion prediction tasks. During the pre-training stage, various masking strategies are applied to the agent and map polylines, and the encoder is trained to reconstruct the missing data. In the fine-tuning stage, the reconstruction heads are removed and the encoder is frozen; the decoder is then trained with complete data for the specific prediction task. However, this can cause the decoder to under-perform when faced with incomplete historical trajectories despite the encoder's exposure to such data during pre-training.

For feature distillation method\cite{monti2022observationsenoughknowledgedistillation}, a well-trained teacher model using complete input trajectories guides a student mo\begin{figure*}[!t]
    \centering
    \includegraphics[width=16cm]{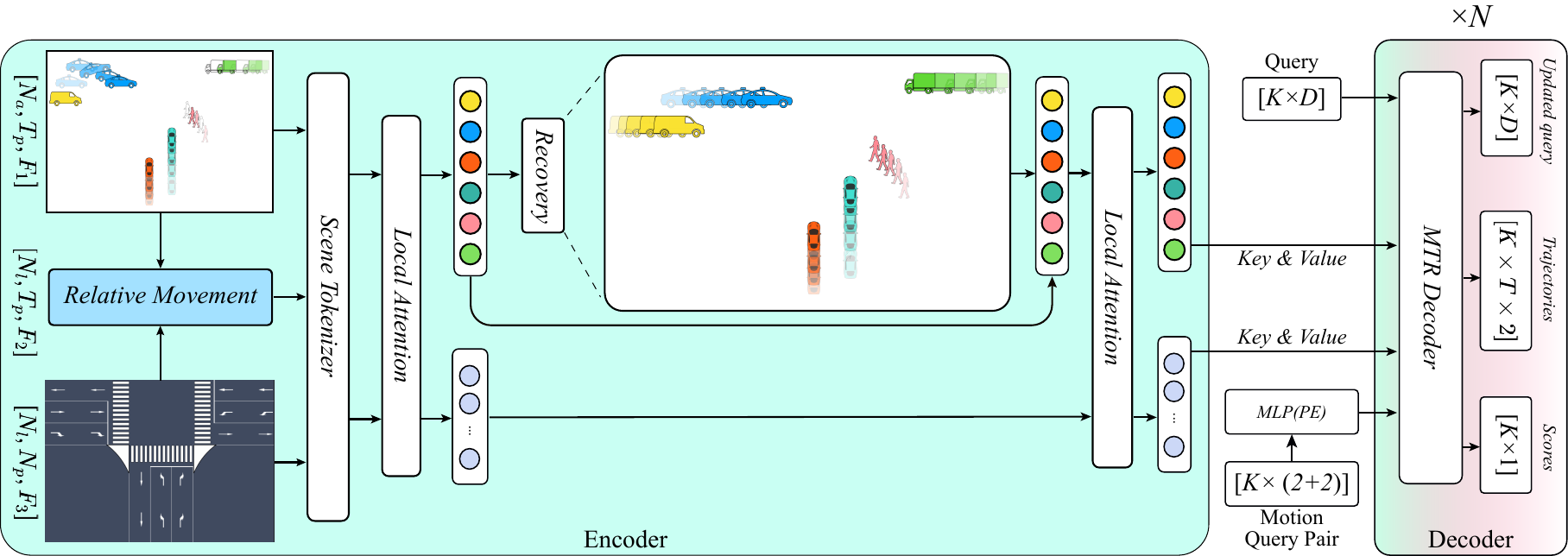} 
    \caption{Overview of the proposed pipeline. The history recovery is conducted in the early stage of the encoder. Compared to previous pipelines, our proposed framework has a more concise structure and can be trained in an end-to-end fashion.}
    \label{fig:encoder-structure}
    \vspace{-1em}
\end{figure*}del using incomplete input trajectories. POP\cite{POP_RAL} combines both methods to address partially observable prediction. While effective, this approach is complex, computationally expensive, and relies heavily on the quality of the teacher model. Some studies \cite{adversarialbackdoorattacknaturalistic, On_adversarial_robustness} have investigated enhancing the robustness of predictors against input noise from the perspective of data poisoning. However, these approaches often overlook the impact of incomplete trajectories. Unlike previous methods, we propose a unified framework that explicitly recovers clean historical trajectories in the encoder stage and integrates this new information back into the agent tokens for further information fusion. As a result, our approach is robust to both noisy and incomplete data while maintaining excellent prediction performance. Additionally, our proposed recovery module is lightweight and plug-and-play. 
\section{Methodology}
\subsection{Problem Formulation and Input Representation}
Following MTR, we adopt a vectorized representation for both maps and agents. And our focus is on marginal motion prediction.  
The historical trajectories of ${N_{a}}$ traffic participants are denoted as $\mathcal{A}=\{a_{1},a_{2},\dots,a_{N_{a}}\}$, where each agent $a_{i}\in \mathbb{R}^{ T_{p} \times F_{1}}$ has $T_{p}$ past timesteps and $F_{1}$ feature dimensions. The corresponding map is partitioned into $N_{l}$ polylines $\mathcal{M}=\{m_{1},m_{2},\dots,m_{N_{l}}\}$, with each polyline $m_{i}\in \mathbb{R}^{ N_{p} \times F_{2}}$ comprising $N_{p}$ points and $F_{2}$ feature dimensions. The predictor will anticipate $K$ different modality future trajectories $\mathcal{Z}=\{z_{1},z_{2},\dots,z_{N_{a}}\}$ over the future $T_{f}$ timesteps, where $z_{i}=\{z_{i}^{1},z_{i}^{2},\dots,z_{i}^{K}\}\in \mathbb{R}^{ K \times T_{f} \times D}$. The confidence score for $z_{i}$ are denoted as $p_{i} = \{p_{i}^{1},p_{i}^{2},\dots,p_{i}^{K}\}$.
Then for selected agent $z_{i}$, existing motion prediction task aims to estimate the distribution: 
\begin{equation}
P(z_{i} \mid \mathcal{M}, \mathcal{A}) = \sum_{k=1}^{K} p_k^i P(z_{i}^{k} \mid \mathcal{M},\mathcal{A})
\end{equation}

Different with other frameworks, we prioritize reconstructing a clean and complete historical trajectory $a_i^{re}$ in the whole prediction pipeline. Besides, we incorporate relative historical movements $\mathcal{R}$ between selected agent $a_i$ and
map polylines $\mathcal{M}$ as additional context input, which can be easily calculated by matrix operation. We use $\mathcal{V} = \{v_{1},v_{2},\dots,v_{N_{a}}\}\in\mathbb{R}^{N_{a}\times T_{p}\times 1}$ to denote the validity of each timestep observation. Hence, we rewrite the probability distribution as:

\begin{equation}
\label{eq-probability}
\begin{aligned}
&P(z_{i},a_i^{re} \mid \mathcal{A}*\mathcal{V},\mathcal{M} ,\mathcal{R}) \\
&= \sum_{k=1}^{K} p_k^i P(z_{i}^{k},a_i^{re} \mid \mathcal{A}*\mathcal{V},\mathcal{M},\mathcal{R}) \\
&= \sum_{k=1}^{K} p_k^i P(z_{i}^{k} \mid a_i^{re},\mathcal{A}*\mathcal{V},\mathcal{M},\mathcal{R}) \cdot P(a_i^{re} \mid \mathcal{M},\mathcal{A}*\mathcal{V},\mathcal{R})
\end{aligned}
\end{equation}

We claim that local agents and map structures surrounding selected agent are sufficient to reconstruct historical trajectories. Since $\mathcal{R}$ depends on $\mathcal{M},\mathcal{A}$, Eq. \ref{eq-probability} can be further simplified as:

\begin{equation}
\begin{aligned}
&P(z_{i},a_i^{re} \mid \mathcal{A}*\mathcal{V},\mathcal{M},\mathcal{R} ) 
=P(z_{i},a_i^{re} \mid \mathcal{A}*\mathcal{V},\mathcal{M})\\
&\approx \sum_{k=1}^{K} p_k^i P(z_{i}^{k} \mid a_i^{re},\mathcal{A}*\mathcal{V},\mathcal{M}) \cdot P(a_i^{re} \mid local(\mathcal{A}*\mathcal{V},\mathcal{M}))
\end{aligned}
\end{equation}

 By employing an agent-centric strategy, the input for each selected agent can be represented as follows:
1) Agent historical information $\mathcal{A} \in \mathbb{R}^{N_{a} \times T_{p} \times F_{1}}$ where \(F_{1}\) includes features such as position, heading, velocity, acceleration, agent type, agent size, valid sign, and one-hot embeddings for past timesteps.
2) Map context information
$\mathcal{M} \in \mathbb{R}^{N_{l} \times N_{p} \times F_{2}}$,
where \(F_{2}\) includes position, direction, and waypoint type. 3) Relative historical movements between selected agents and maps
$\mathcal{R} \in \mathbb{R}^{N_{l} \times T_{p} \times F_{3}}$,
where \(F_{3}\) indicates the relative position and orientation between the selected agent and the center of each road polyline over the past $T_{p}$ timesteps.

\subsection{Network Encoder}
\begin{figure}[ht]
    \centering
    \includegraphics[width=1\linewidth]{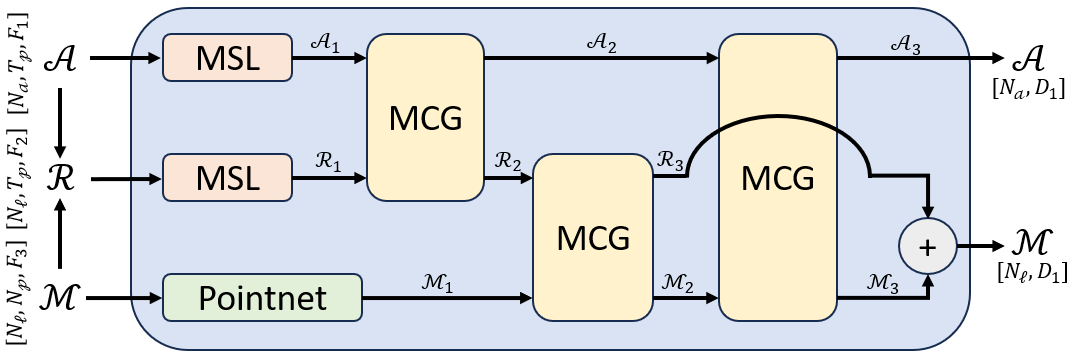}
    \caption{Detailed design of the proposed scene tokenization module.}
    \label{fig:scene_tokenization}
    \vspace{-1em}
\end{figure}

\textbf{Scene Tokenization.}
Temporal information, including agent historical states $\mathcal{A}$ and relative movement $\mathcal{R}$, is processed using a Multi-Scale LSTM (MSL) model. Initially, the data is concurrently passed through a 1D CNN module with kernel sizes of $1$, $3$, and $5$. It then progresses through a two-layer LSTM network, where the output at the final timestep is captured, concatenated across feature dimensions, and passed through an additional MLP layer to generate the final feature token (see Fig. \ref{fig:scene_tokenization}). For the road polylines data $\mathcal{M}$, a simple PointNet-like network is employed to extract the spatial features of each polyline.
\begin{equation}
\begin{array}{clcl}
    \mathcal{A}_{1} &= \text{MSL}(\mathcal{A}), &\mathcal{A}_{1} \in \mathbb{R}^{N_{a}\times D_{1}}, \\
    \mathcal{R}_{1} &= \text{MSL}(\mathcal{R}), &\mathcal{R}_{1} \in \mathbb{R}^{N_{l}\times D_{1}}, \\
    \mathcal{M}_{1} &= \phi(\text{MLP}(\mathcal{M})), &\mathcal{M}_{1}\in \mathbb{R}^{N_{l}\times D_{1}},
\end{array}
\end{equation}

where $\phi(\cdot)$ denotes the max-pooling operation.
In this process, we integrate encodings from various input modalities using Multi-Context Gating (MCG) as proposed in \cite{MultiPath++}. We utilize a cascading method where, at each stage, two distinct modalities are chosen from a set of three to be input into the MCG module. The output from one MCG module is then fed into the subsequent MCG module in the sequence, as illustrated in Fig.\ref{fig:scene_tokenization}.

\begin{equation}
\begin{split}
    (\mathcal{A}_{2}, \mathcal{R}_{2}) &= \text{MCG}(\mathcal{A}_{1}, \mathcal{R}_{1}), \\
    (\mathcal{M}_{2}, \mathcal{R}_{3}) &= \text{MCG}(\mathcal{M}_{1}, \mathcal{R}_{2}), \\
    (\mathcal{A}_{3}, \mathcal{M}_{3}) &= \text{MCG}(\mathcal{A}_{2}, \mathcal{M}_{2}).
\end{split}  
\label{eqation:mcg}
\end{equation}
Then, we use $\mathcal{A}_{3}$ as the final agent tokens $\mathcal{A}_{agent}\in \mathbb{R}^{N_{a}\times D_{1}}$ and $\mathcal{M}_{3}+\mathcal{R}_{3}$ as the final map tokens $\mathcal{M}_{map}\in \mathbb{R}^{N_{p}\times D_{1}}$.

We assume that local interactions are crucial and sufficient to reconstruct incomplete historical trajectories. Therefore, we use a single layer of local attention to ensure that each agent token attends to its K nearest neighbor tokens (including both agent and map tokens). Following this, a simple MLP layer is employed to recover the complete historical trajectories for all agents. Next, a PointNet-like layer aggregates these recovered trajectories into agent tokens. A residual connection is added between the input and output of the recovery module to ensure stable training (see Fig.\ref{fig:encoder-structure}).
\begin{equation}
    \begin{split}
       \mathcal{A}_{agent} = \mathcal{A}_{agent} + \text{Recovery}(\mathcal{A}_{agent}),\\
        \text{Recovery}(\mathcal{A}_{agent}) =\text{MLP}(\mathcal{A}_{Past}),\\
        \mathcal{A}_{Past} = \text{MLP}(\mathcal{A}_{agent}),
    \end{split}
\end{equation}
Where $\mathcal{A}_{Past} \in \mathbb{R}^{N_{a}\times ({T_{p} \times 4})}$ and $4$ denotes the position and velocity for the recovered historical trajectories.
This recovery module aims to reconstruct the incomplete historical trajectories for each agent and integrate this enriched historical information into the agent tokens. After the recovery module, both agent tokens and map tokens will go through another four layers of local attention for further feature fusion.
The $i_{th}$ transformer encoder layer can be formulated as:
\begin{equation}
    \begin{split}
        Q^{i} = \text{MHA} (
            & Q^{i-1}+\text{PE}({Q^{i-1}}), \\ 
            & \mathcal{K}(Q^{i-1}) + \text{PE}({\mathcal{K}(Q^{i-1}))}, 
            \mathcal{K}(Q^{i-1}) 
        ),
    \end{split}
\end{equation}

where $\text{MHA}(\cdot,\cdot,\cdot)$ represents the multi-head attention function, $Q^{0}=[\mathcal{M}_{map},\mathcal{A}_{agent}]\in \mathbb{R}^{(N_{a}+N_{m})\times D_{1}}$, and
$\mathcal{K}(\cdot)$ denotes the $K$-nearest neighbours (KNN) algorithm, which is used to identify the $K$ nearest tokens relative to each query. The term $\text{PE}(\cdot)$ refers to the sinusoidal positional encoding assigned to input tokens, incorporating the most recent position of each agent and the central point of each map polyline. The final output of the local attention layer will be sent into the MTR decoder, $[\mathcal{M}_{map},\mathcal{A}_{agent}]=Q^{Final}$.
\subsection{Network Decoder}
The decoder architecture is identical to the MTR decoder from \cite{shi2023MTR}, except for the method used to calculate the loss from the output trajectories. Between the output of each decoder layer and the loss calculation, we apply evolving and distinct anchor techniques as described in \cite{lin2024eda}. We use 6 decoder layers, with evolving anchors applied at the second and fourth layers, and distinct anchors selected at each layer. 

\subsection{Loss Function}
We put forward a combined loss function which is consist of two components: Original MTR\cite{shi2023MTR} loss and recovery loss.  The total loss function can be defined as follows:\\
\begin{equation}
\mathcal{L}_{Total} = \mathcal{L}_{MTR} + \mathcal{L}_{Recovery} 
\end{equation}
For the MTR loss, We follow the loss function of MTR\cite{shi2023MTR}, using a decoder loss $\mathcal{L}_{Decoder}$ and a dense future prediction loss $\mathcal{L}_{Df}$. 

The recovery loss $\mathcal{L}_{Recovery}$ aims to optimize the recovery module to resume the incomplete historical trajectories and it is simply the $\mathcal{L}_{1}$ loss of recovered $\mathcal{A}_{Past}$ and ground truth historical trajectories $\mathcal{A}_{PastGT}$.

\section{Experiments}

\begin{table*}[htbp]
\centering
\caption{Prediction on the test leaderboard of the motion prediction track of the Waymo Open Dataset Challenge. Our approach is termed RMP, i.e., Robust Motion Predictor. Soft mAP is the official ranking metric, while miss rate is the secondary ranking metric. The first place is denoted by \textbf{bold}, the second place by \underline{underline}, and the third place by \textasteriskcentered{asterisk}.}
\label{tab:leadboard}
\begin{tabular}{cccccccc} 
    \hline
    \\[-1em]
      \rowcolor[rgb]{0.863,0.863,0.863} Waymo Competition ~& Method & Soft mAP $\uparrow$ & mAP $\uparrow$ & minADE $\downarrow$ & minFDE $\downarrow$ & Miss Rate $\downarrow$ & Overlap Rate $\downarrow$\\
    \hline
    \\[-1em]
    \multirow{13}{*}{2024}& MTR v3\cite{shi2024mtrv3} & \textbf{0.4967} &  \textbf{0.4859} & \underline{0.5554} & \underline{1.1062} &  \underline{0.1098} & 0.1279 \\
    & ModeSeq\cite{zhou2024modeseqtamingsparsemultimodal} & \underline{0.4737} & \underline{0.4665} & $0.5680^{*}$ & 1.1766 & 0.1204 & 0.1275 \\
    & Betop\cite{liu2024reasoningmultiagentbehavioraltopology} & 0.4698 & $0.4587^{*}$ & 0.5716 & 1.1668 & 0.1183 & 0.1272 \\
    & BehaveOcc & 0.4678 & 0.4566 & 0.5723 & 1.1668 & 0.1176 & 0.1278 \\
    & QMTR & 0.4649 & 0.4445 & 0.5702 & 1.1627 & 0.1177 & 0.1269 \\
    & EDA\cite{lin2024eda} & 0.4596 & 0.4487 & 0.5718 & 1.1702 & 0.1169 & $0.1266^{*}$ \\
    & ControlMTR\cite{sun2024controlmtrcontrolguidedmotiontransformer} & 0.4572 & 0.4414 & 0.5897 & 1.1916 & 0.1282 & \textbf{0.1259} \\
    & LLM-Augmented-MTR\cite{llm_augmented_mtr} & 0.4423 & 0.4270 & 0.5987 & 1.2084 & 0.1316 & 0.1274 \\
    & MTR\cite{shi2023MTR} & 0.4403 & 0.4249 & 0.5964 & 1.2039 & 0.1312 & 0.1274 \\
    &{\cellcolor[rgb]{0.902,0.902,0.902}}RMP Ensemble & {\cellcolor[rgb]{0.902,0.902,0.902}}\underline{0.4737} & {\cellcolor[rgb]{0.902,0.902,0.902}}0.4531 & {\cellcolor[rgb]{0.902,0.902,0.902}}0.5564 & {\cellcolor[rgb]{0.902,0.902,0.902}}$1.1188^{*}$ & {\cellcolor[rgb]{0.902,0.902,0.902}}\textbf{0.1084} & {\cellcolor[rgb]{0.902,0.902,0.902}} \textbf{0.1259} \\
    &{\cellcolor[rgb]{0.902,0.902,0.902}}RMP & {\cellcolor[rgb]{0.902,0.902,0.902}}0.4673 & {\cellcolor[rgb]{0.902,0.902,0.902}}0.4523 & {\cellcolor[rgb]{0.902,0.902,0.902}}0.5739 & {\cellcolor[rgb]{0.902,0.902,0.902}}1.1698 & {\cellcolor[rgb]{0.902,0.902,0.902}}$0.1159^{*}$ & {\cellcolor[rgb]{0.902,0.902,0.902}} 0.1266 \\
    &{\cellcolor[rgb]{0.902,0.902,0.902}}RMP e2e & {\cellcolor[rgb]{0.902,0.902,0.902}}0.3828 & {\cellcolor[rgb]{0.902,0.902,0.902}}0.3440 & {\cellcolor[rgb]{0.902,0.902,0.902}}\textbf{0.5529} & {\cellcolor[rgb]{0.902,0.902,0.902}}\textbf{1.0932} & {\cellcolor[rgb]{0.902,0.902,0.902}}0.1354 & {\cellcolor[rgb]{0.902,0.902,0.902}}0.1295 \\
    \hline
    \\[-1em]
    \multirow{6}{*}{Previous Years}
    & DenseTNT\cite{gu2021densetnt} & - & 0.3281 & 1.0387 & 1.5514 & 0.1573 & 0.1779 \\
    & SceneTransformer\cite{ngiam2022scenetransformer} & - & 0.2788 & 0.6117 & 1.2116 & 0.1564 & 0.1473 \\
    & ReCoAt\cite{huang2022recoatdeeplearningbasedframework} & - & 0.2711 & 0.7703 & 1.6668 & 0.2437 & 0.1642 \\
    & HDGT\cite{JiaHDGT} & 0.3709 & 0.3577 & 0.5933 & 1.2055 & 0.1511 & 0.1557 \\
    & MoST\cite{mu2024mostmultimodalityscenetokenization} & 0.4396 & 0.4201 & 0.5391 & 1.1099 & 0.1172 & - \\
    & MTR++\cite{shi2024mtr++} & 0.4410 & 0.4329 & 0.5906 & 1.1939 & 0.1298 & 0.1281 \\
    & MGTR\cite{gan2024mgtrmultigranulartransformermotion} & 0.4599 & 0.4505 & 0.5918 & 1.2135 & 0.1298 & 0.1275 \\
    \hline
\end{tabular}
\end{table*}

\begin{figure*}[h]
    \centering
    \includegraphics[width=0.95\linewidth]{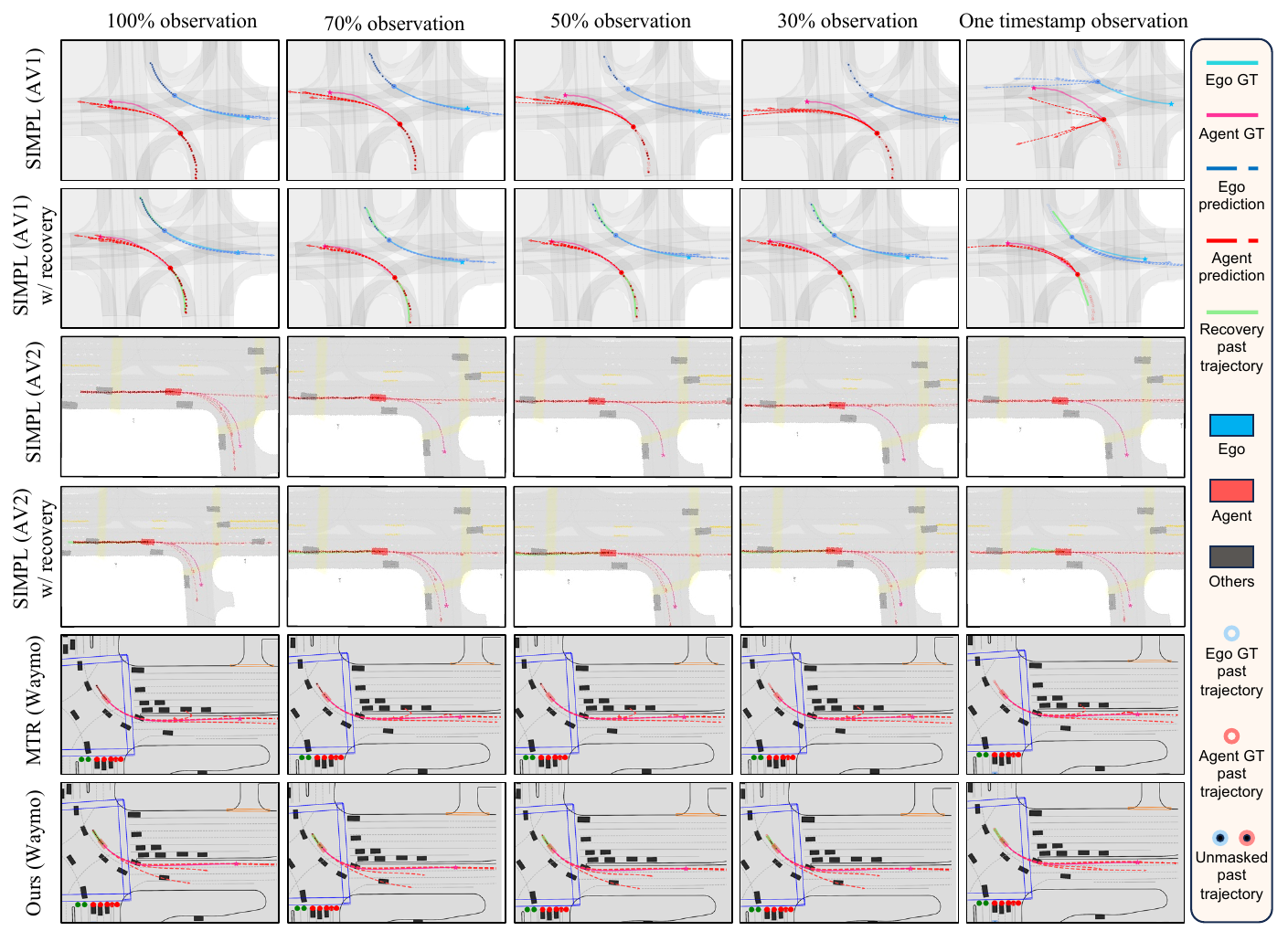}
    \caption{Visualization result of our proposed recovery module and prediction results under different observation ratios.}
    \label{fig:simpl_av1}
\end{figure*}

\subsection{Experimental Setup}
\subsubsection{Dataset and Metrics}
Our RMP-YOLO model is trained using the Waymo Open Motion Dataset (WOMD), which consists of 486,995 scenes for training and 44,097 scenes for validation. Additionally, we evaluate our proposed recovery module on other state-of-the-art (SOTA) prediction methods based on the Argoverse 1\&2  datasets. For WOMD, we use Soft mAP as our key evaluation metric, while for Argoverse 1\&2, we select Brier-FDE$_{k}$ as the main metric. Additional metrics such as minADE, minFDE, Miss Rate, and Overlap Rate are also used to provide supplementary evaluation of the models.

\subsubsection{Training Details} 
We use the AdamW optimizer to train our model in an end-to-end manner, with an initial learning rate set to 1.0e-4. Beginning at epoch 20, the learning rate is halved every two epochs. We train the model for 30 epochs and then fine-tune it for an additional 10 epochs, maintaining a learning rate of 6.25e-6. We train our models on 4 Nvidia-A6000 GPUS with total 48 batch size. We randomly mask 70\% input data and our proposed recovery module is supervised to reconstruct the complete input data. 
\subsection{Leaderboard Performance}
\begin{table*}[!t]
\centering
\caption{Ablation Study on the Scene Tokenization and Recovery Modules.}
\label{table-ablation}
\setlength{\extrarowheight}{0pt}
\setlength{\aboverulesep}{0pt}
\setlength{\belowrulesep}{0pt} 
\resizebox{0.9\linewidth}{!}{%
\begin{tabular}{cccccccccc} 
\toprule
\rowcolor[rgb]{0.863,0.863,0.863} Baseline(MTR) & \begin{tabular}[c]{@{}>{\cellcolor[rgb]{0.863,0.863,0.863}}c@{}}w/ Scene \\Tokenization\end{tabular} & \begin{tabular}[c]{@{}>{\cellcolor[rgb]{0.863,0.863,0.863}}c@{}}w/ Recovery \\(RM 70\%)\end{tabular} & \begin{tabular}[c]{@{}>{\cellcolor[rgb]{0.863,0.863,0.863}}c@{}}w/ Recovery \\(no RM)\end{tabular} & Soft mAP $\uparrow$ & mAP $\uparrow$ & minADE $\downarrow$ & minFDE $\downarrow$ & Miss Rate $\downarrow$ & Overlap Rate $\downarrow$ \\ 
\hline
\checkmark &  &  &  & 0.3616 & 0.3494 & 0.6737 & 1.3725 & 0.1654 & \textbf{0.1341} \\
\checkmark & \checkmark &  &  & 0.4033 & 0.3889 & \textbf{0.6310} & \textbf{1.3104} & 0.1468 & 0.1346 \\
\checkmark & \checkmark & \checkmark &  & 0.3992 & 0.3841 & 0.6357 & 1.3179 & \textbf{0.1450} & 0.1367 \\
\checkmark & \checkmark &  & \checkmark & \textbf{0.4046} & \textbf{0.3903} & 0.6340 & 1.3203 & 0.1478 & 0.1359 \\
\bottomrule
\end{tabular}
}
\vspace{-1em}
\end{table*}

\subsubsection{Leaderboard}
As shown in TABLE \ref{tab:leadboard}, our method, RMP, achieves competitive results, with our ensemble version securing second place in Soft mAP (0.4737) and achieving the best miss rate (0.1084) and overlap rate (0.1259). The top-performing model in terms of Soft mAP is MTR v3, with a score of 0.4967. Additionally, MTR v3 also leads in minADE (0.5554) and minFDE (1.1062), while our method's variant RMP\_e2e achieves the lowest minADE (0.5529) and minFDE (1.0932). RMP\_e2e is trained in an end-to-end fashion to directly generate 6 futures without using Non-Maximum Suppression. These results demonstrate the SOTA performance of our RMP approach across various evaluation metrics. The real-world deployment video can be viewed in the submission file.
\begin{figure}[h]
    \centering
    \includegraphics[width=0.8\linewidth]{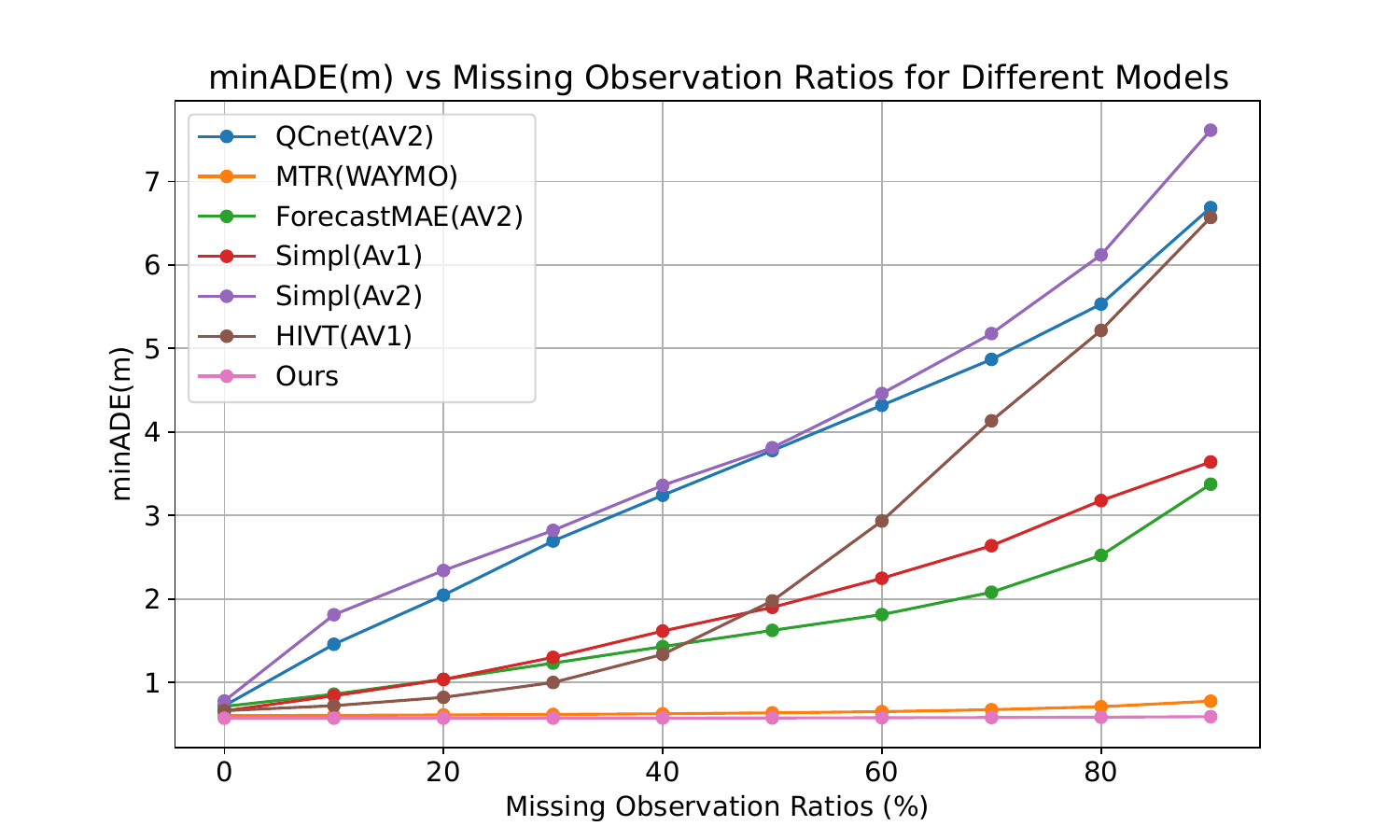}
    \caption{Performance of various methods using deficient historical trajectory data.}
    \label{fig:all_comparison}
    \vspace{-2em}
\end{figure}
\subsection{Robustness}
\begin{figure}[h]
    \centering
    \includegraphics[width=1 \linewidth]{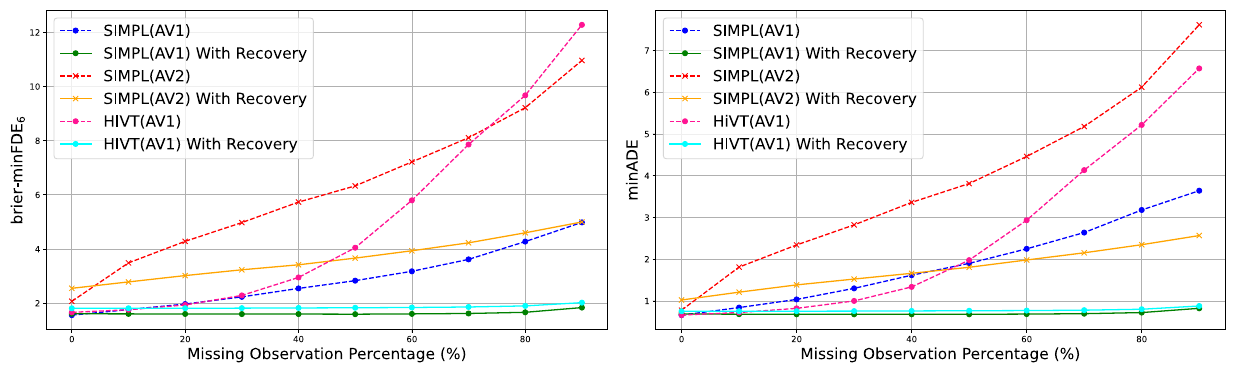}
    \caption{Performance of model using deficient historical trajectory data (Argoverse 1 \& 2).}
    \label{fig:data_1}
    \vspace{-1em}
\end{figure}

\begin{figure}[h]
    \centering
    \includegraphics[width=1 \linewidth]{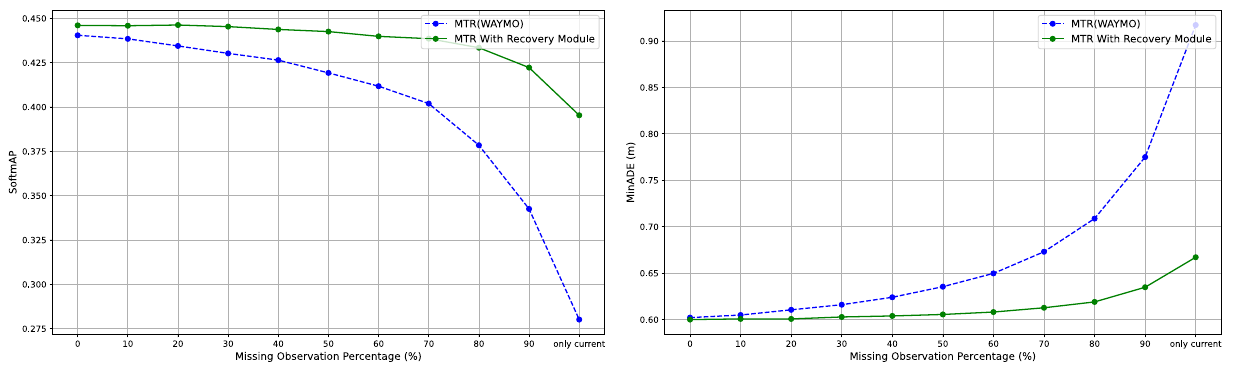}
    \caption{Performance of model using deficient historical trajectory data (Waymo).}
    \label{fig:data_2}
    \vspace{-1.2em}
\end{figure}

Unlike other models(see Fig.\ref{fig:all_comparison}), which exhibit a significant degradation in accuracy as the missing observation ratios increase, our approach proves to be more robust and reliable, particularly in challenging scenarios with higher levels of data loss.
The inclusion of the recovery module demonstrates a clear improvement in performance compared to models like SIMPL and HIVT which lack this feature. When the recovery method is applied, these models consistently achieve lower brier-minFDE and minADE values, even as the percentage of missing observations increases (see Fig. \ref{fig:data_1}). We also performed the same study using MTR as the baseline model on the Waymo motion dataset, evaluating the SoftmAP and minADE metrics (see Fig. \ref{fig:data_2}). This indicates that our approach effectively handles data loss, resulting in more reliable and stable predictions even in highly uncertain environments. Please check out our submission video (due to page limitation) for qualitative results demonstrating the robustness of our method against noisy input.

\begin{figure}[h]
    \centering
    \includegraphics[width=1 \linewidth]{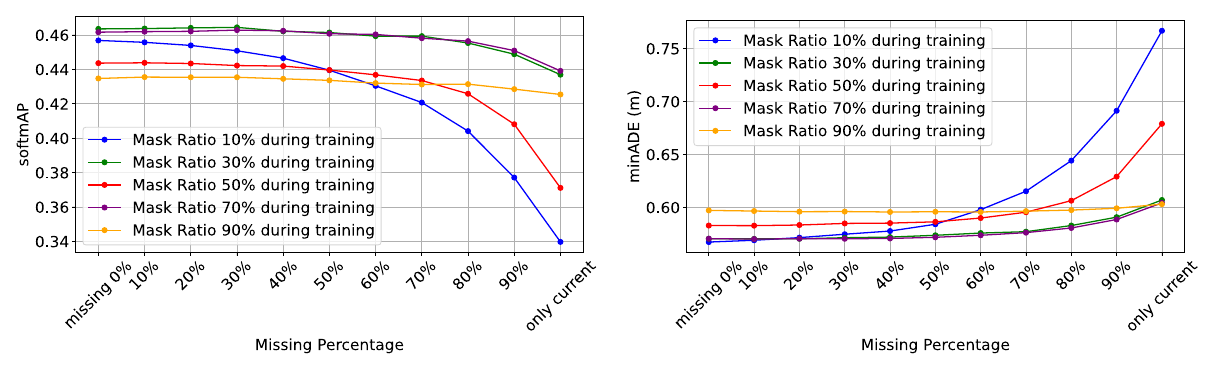}
    \caption{Study on the effect of training with different masking ratios on the inference performance and robustness.}
    \label{fig:masking-ratios}
    \vspace{-1.4em}
\end{figure}

\subsection{Ablation Study}
To explore the impact of different mask ratios on model performance, we conducted an ablation study, as shown in Fig.\ref{fig:masking-ratios}. Based on the results, we selected a mask ratio of 0.7 to achieve the best balance between robustness and prediction accuracy. In order to quantitatively assess the impact of our proposed scene tokenization and recovery modules, we conducted a series of ablation studies(on 20\% training dataset, 100\%validtion dataset) and summarized the results in TABLE \ref{table-ablation}. The comparison with the baseline shows significant improvements in the model's Soft mAP performance, indicating the effectiveness of the proposed modules.

\subsection{Inference Efficiency}
We integrated our recovery module into established methods to assess its effect on the model's inference efficiency. The results are documented in TABLE \ref{table-efficiency}. By comparing the number of parameters and the model's runtime before and after adding our recovery module, we can conclude that our proposed recovery module is highly lightweight and minimally impacts the model's efficiency.

\begin{table}[h]
\centering
\caption{Inference Efficiency of the Recovery Module}
\label{table-efficiency}
\setlength{\extrarowheight}{0pt}
\addtolength{\extrarowheight}{\aboverulesep}
\addtolength{\extrarowheight}{\belowrulesep}
\setlength{\aboverulesep}{0pt}
\setlength{\belowrulesep}{0pt}
\resizebox{0.99\linewidth}{!}{%
\begin{tabular}{ccccc} 
\toprule
\rowcolor[rgb]{0.863,0.863,0.863} {\cellcolor[rgb]{0.863,0.863,0.863}} & \multicolumn{2}{c}{Baseline} & \multicolumn{2}{c}{w/ Recovery Module} \\
\rowcolor[rgb]{0.863,0.863,0.863} \multirow{-2}{*}{{\cellcolor[rgb]{0.863,0.863,0.863}}Method} & Parameters (M) & Run Time (ms) & Parameters (M) & Run Time (ms) \\ 
\hline
SIMPL (AV2) & 2.65 & 48.75 & 2.84 & 61.88 \\
MTR & 65.78 & 80.20 & 66.55 & 81.10 \\
Ours & 68.68 & 100.23 & 69.45 & 100.89 \\
\bottomrule
\end{tabular}
}
\vspace{-2em}
\end{table}

\section{CONCLUSIONS}
We introduced RMP-YOLO, a motion prediction framework designed to effectively manage incomplete historical trajectory inputs. The framework prioritizes reconstructing full past trajectories by leveraging local map topology and agent interactions. Extensive experiments demonstrate our model's competitive prediction performance and robustness against incomplete input trajectories. In the future, we plan to upgrade our framework to a query-centric paradigm to further enhance inference speed.

\addtolength{\textheight}{-1cm}   
\bibliographystyle{IEEEtran}
\bibliography{root}

\end{document}